%%%%%%%%%%%% MACROS FOR PAPERS AND REPORTS %%%%%%%%%%%%
%My little macros

\def\ignore#1{}
 
%%%%%%%%%%%%%%%%%%%%%

\newcount\sectnum
\newcount\subsectnum
\newcount\eqnumber

\global\eqnumber=1\sectnum=0

% Equation labels

\def\lab{(\the\sectnum.\the\eqnumber)}

%Example of use: suppose we want to give a label \lgh to an equation
% $$ ......  \xdef\lgh{\lab} \eqnum \show{lgh}$$
% Later refer to Eq. \lgh\ ...
% Note the \ after \lgh; it seems to be needed if we want the equation number
% to be followed by a space; not needed if followed by . or ,

%The next macro is used to display labels in drafts, so that you do
%not have to remember them

\def\show#1{}

%The next macro is to be used for final drafts that do not display labels
%\def\show#1{}

%%%%%%%%%%%%%%%%%%%%%

\def\smskip{\vskip 5 pt}
\def\medskip{\vskip 10 pt}
\def\bigskip{\vskip 15 pt}
\def\pn{\par\noindent}

\def\frac#1#2{{#1\over #2}}

\def\m{\mu}
\def\p{\pi}

\def\re{\Re}
\def\rn{\Re^n}

 %break line; horizontal space
\def\tl{\tilde}

\def\old#1{}% invalidates text in braces 
\def\leaderfill{\leaders\hbox to 1em{\hss.\hss}\hfill}
%Example of use: \line{1. Optimality Conditions\leaderfill p.\ 2}

% John's macros

\parindent=2pc
\baselineskip=15pt
\vsize=8.7 true in
\voffset=0.125 true in
\parskip=3pt

% vector/matrix macros

%eqalign macros
\def\minprob#1#2#3{$$\eqalign{&\hbox{minimize\ \ }#1\cr &\hbox{subject to\ \
}#2\cr}\ifnum 0=#3{}\else\eqno(#3)\fi$$}        
     
\def\maxprob#1#2#3{$$\eqalign{&\hbox{maximize\ \ }#1\cr &\hbox{subject to\ \
}#2\cr}\ifnum 0=#3{}\else\eqno(#3)\fi$$}        
     
\def\aligntwo#1#2#3#4#5{$$\eqalign{#1&#2\cr #3&#4\cr}
\ifnum 0=#5{}\else\eqno(#5)\fi$$}
\def\alignthree#1#2#3#4#5#6#7{$$\eqalign{#1&#2\cr #3&#4\cr #5&#6\cr}
\ifnum 0=#7{}\else\eqno(#7)\fi$$}

% Macros to automatically advance equation and other numbers

\def\eqnum{\eqno{\hbox{(\the\sectnum.\the\eqnumber)}\global\advance\eqnumber
by1}}

\def\eqnu{\eqno{\hbox{(\the\sectnum.\the\eqnumber)}\global\advance\eqnumber
by1}}

\newcount\examplnumber
\def\examplnum{\global\advance\examplnumber by1}

\newcount\figrnumber
\def\figrnum{\global\advance\figrnumber by1}

\newcount\propnumber
\def\propnum{\global\advance\propnumber by1}

\newcount\defnumber
\def\defnum{\global\advance\defnumber by1}

\newcount\lemmanumber
\def\lemmanum{\global\advance\lemmanumber by1}

\newcount\assumptionnumber
\def\assumptionnum{\global\advance\assumptionnumber by1}

\newcount\conditionnumber
\def\conditionnum{\global\advance\conditionnumber by1}

\def\exampl{\the\sectnum.\the\examplnumber}
\def\figr{\the\sectnum.\the\figrnumber}
\def\propn{\the\sectnum.\the\propnumber}
\def\defn{\the\sectnum.\the\defnumber}
\def\lemman{\the\sectnum.\the\lemmanumber}
\def\assumptionn{\the\sectnum.\the\assumptionnumber}
\def\condn{\the\sectnum.\the\conditionnumber}

\def\section#1{\goodbreak\vskip 3pc plus 6pt minus 3pt\leftskip=-2pc
   \global\advance\sectnum by 1\eqnumber=1
\global\examplnumber=1\figrnumber=1\propnumber=1\defnumber=1\lemmanumber=1\assumptionnumber=1 \conditionnumber =1 \subsectnum=0%
   \line{\hfuzz=1pc{\hbox to 3pc{\bf %\the\sectnum.\quad
   \vtop{\hfuzz=1pc\hsize=38pc\hyphenpenalty=10000\noindent\uppercase{\the\sectnum.\quad #1}}\hss}}
			\hfill}
			\leftskip=0pc\nobreak\tenf
			\vskip 1pc plus 4pt minus 2pt\noindent\ignorespaces}

% ETP Macros

%\def\section#1{\goodbreak\vskip 3pc plus 6pt minus 3pt\leftskip=-2pc
%   \global\advance\sectnum by 1\eqnumber=1
%   \line{\hfuzz=1pc{\hbox to 3pc{\bf %\the\sectnum.\quad
%   \vtop{\hfuzz=1pc\hsize=38pc\hyphenpenalty=10000\noindent\uppercase{#1}}\hss}}
%                        \hfill}
%                        \leftskip=0pc\nobreak\tenf
%                        \vskip 1pc plus 4pt minus 2pt\noindent\ignorespaces}

\def\sect#1{\noindent\leftskip=-2pc\tenf
   \goodbreak\vskip 1pc plus 4pt minus 2pt
                \global\advance\subsectnum by 1\eqnumber=1
   \line{\hfuzz=1pc{\hbox to 3pc{\bf %\the\sectnum.\quad
   \vtop{\hfuzz=1pc\hsize=38pc\hyphenpenalty=10000\noindent\uppercase{{\bf #1}}}\hss}}
                        \hfill}
   \leftskip=0pc\nobreak\tenf
                        \vskip 1pc plus 4pt minus 2pt\nobreak\noindent\ignorespaces}

\def\subsection#1{\noindent%
   \goodbreak\vskip 1pc plus 4pt minus 2pt%
 		\global\advance\subsectnum by 1%
   \line{\hfuzz=1pc{\hbox to 3pc%
   {\bf  \vtop{\hfuzz=1pc\hsize=38pc\hyphenpenalty=10000\noindent{\bf 
  \the\sectnum.\the\subsectnum\ \ \ #1}}\hss}}%
			\hfill}%
   \nobreak%
			\vskip 1pc plus 4pt minus 2pt\nobreak\noindent\ignorespaces}%

\def\subsubsection#1{\goodbreak\vskip 1pc plus 4pt minus 2pt
   \hfuzz=3pc\leftskip=0pc\noindent\tenit #1 \nobreak\tenf\vskip 6pt plus 1pt
                                minus 1pt\nobreak\ignorespaces\leftskip=0pc}
%
%\def\rthl{Sec. \the\chapnum.\the\sectnum}                      
%\def\rthc{#1}\nobreak\noindent\ignorespaces
%\newcount\sectnum \sectnum=0
%\newcount\subsectnum \subsectnum=0

\def\beginexample#1{\noindent\goodbreak\vskip 6pt plus 1pt minus 1pt
\noindent
  \hbox {\bf Example #1\hss}%\break%\noindent
  \nobreak\vskip 4pt plus 1pt minus 1pt \nobreak\noindent\ninef
  \global\advance
                \leftskip by\parindent\pn}
\def\endexample{\vskip 12pt\tenf\par
  \global\advance\leftskip by -\parindent
  }

\def\beginexercise#1{\noindent\goodbreak\vskip 6pt plus 1pt minus 1pt \noindent\global\normalbaselineskip=12pt
  \hbox {\bf Exercise #1\hss}%\break%\noindent
  \nobreak\vskip 4pt plus 1pt minus 1pt 
  \nobreak\noindent\ninef\global\advance\leftskip
                        by\parindent\pn}
\def\endexercise{\vskip 12pt\tenf\par
  \global\advance\leftskip by -\parindent
  }

\def\beginsection#1{\noindent\goodbreak\vskip 6pt plus 1pt minus 1pt \noindent\global\normalbaselineskip=12pt
  \hbox {\it #1\hss}
  \vskip 0.1pt plus 1pt minus 1pt \nobreak\noindent\ninef\global\advance
                \leftskip by\parindent\noindent\pn}
\def\endsection{\vskip 12pt\tenf\par
  \global\advance\leftskip by -\parindent
}

%

% Header/Title macros

\def\ref{\smskip\pn}

\def\chapter#1#2{{\bf \centerline{\helbigbig
{#1}}}\bigskip\bigskip{\bf \centerline{\helbigbig
{#2}}}\bigskip\bigskip} % ex. \chapter{Chapter 1}{Title of chapter}

 % ex. \longchapter{Chapter 1}{Title of chapter}{Title of
 %chapter}

 % ex. \papertitle{Title of paper}{Names of Authors}

\def\longpapertitle#1#2#3{{\bf \centerline{\helbigb
{#1}}}\bigskip{\bf \centerline{\helbigb
{#2}}}\bigskip\bigskip{\centerline{
by}}\bigskip{\bf \centerline{
{#3}}}\bigskip\bigskip} 
% ex. \longpapertitle{First part of title of paper}
%{2nd part of title of paper}{Names of Authors}

% List macros

\def\nitem#1{\smskip\item{#1}}

\newcount\alphanum
\newcount\romnum

\def\alphaenumerate{\ifcase\alphanum \or (a)\or (b)\or (c)\or (d)\or (e)\or
(f)\or (g)\or (h)\or (i)\or (j)\or (k)\fi}
\def\romenumerate{\ifcase\romnum \or (i)\or (ii)\or (iii)\or (iv)\or (v)\or
(vi)\or (vii)\or (viii)\or (ix)\or (x)\or (xi)\fi}

\def\alist{\begingroup\vskip10pt\alphanum=1% alphabetical list
\parskip=2pt\parindent=0pt \leftskip=3pc
\everypar{\llap{{\rm\alphaenumerate\hskip1em}}\advance\alphanum by1}}

\def\nolist{\begingroup\vskip10pt\alphanum=0% numerical list
\parskip=2pt\parindent=0pt \leftskip=3pc
\everypar{\llap{\global\advance\alphanum by1(\the\alphanum)\hskip1em}}}

\def\romlist{\begingroup\vskip10pt\romnum=1% roman list
\parskip=2pt\parindent=0pt \leftskip=5pc
\everypar{\llap{{\rm\romenumerate\hskip1em}}\advance\romnum by1}}

% romlist indents more than alist or nolist and can be used inside them

%Figure, table, and box macros

\long\def\fig#1#2#3{\vbox{\vskip1pc\vskip#1
\prevdepth=12pt \baselineskip=12pt
\vskip1pc
\hbox to\hsize{\hfill\vtop{\hsize=25pc\noindent{\eightbf Figure #2\ }
{\eightpoint#3}}\hfill}}}%Figure space definition. Example of use:
%\fig{16pc}{1.1}{A network with one central processor and a separate
%communication link to each device.}

\long\def\widefig#1#2#3{\vbox{\vskip1pc\vskip#1
\prevdepth=12pt \baselineskip=12pt
\vskip1pc
\hbox to\hsize{\hfill\vtop{\hsize=28pc\noindent{\eightbf Figure #2\ }
{\eightpoint#3}}\hfill}}}

\long\def\table#1#2{\vbox{\vskip0.5pc
\prevdepth=12pt \baselineskip=12pt
\hbox to\hsize{\hfill\vtop{\hsize=25pc\noindent{\eightbf Table #1\ }
{\eightpoint#2}}\hfill}}}

%Running Head Macros
 
\def\rightheadline#1{\headline{\tenrm\hfil #1}}

% Concept Macros

\long\def\leftfig#1#2{\vbox{\smskip\hsize=220pt
\vtop{{\noindent {\bf #1}}}
\smskip
\noindent
\vbox{{\noindent #2}}
}}

\long\def\rightfig#1#2#3{\vbox{\smskip\vskip#1
\prevdepth=12pt \baselineskip=12pt
\hsize=210pt
\smskip
\vbox{\noindent{\eightbold #2}
\hskip1em{\eightpoint#3}}
}}

\long\def\concept#1#2#3#4#5{\bigskip\hrule
\vbox{\hbox{\leftfig{#1}{#2} \hskip3em
\rightfig{#3}{#4}{#5}} \smskip}
\hrule\bigskip}

% Example of Use: \concept{Title of Concept}{Text}
% {Figure size}{Figure number?}{Figure caption}

\long\def\bconcept#1#2#3#4#5#6#7{
\vbox{
\hbox to \hsize{\vtop{\par #1}}
\concept{#2}{#3}{#4}{#5}{#6}
\hbox to \hsize{\vtop{\par #7}}
\smskip}
}

% Example of Use: \bconcept{Preceding text}{Title of Concept}{Text}
% {Figure size}{Figure number}{Figure caption}{Following text}

% same as concept but without the \hrule's; ready to be boxed

% Put inside a box

\def\boxit#1{\vbox{\hrule\hbox{\vrule\kern3pt
                                \vbox{\kern3pt#1\kern3pt}\kern3pt\vrule}\hrule}}
% example of use: \setbox0=\vbox{.... }; \boxit{\box0}
\def\centerboxit#1{$$\vbox{\hrule\hbox{\vrule\kern3pt
                                \vbox{\kern3pt#1\kern3pt}\kern3pt\vrule}\hrule}$$}
% example of use: \setbox0=\vbox{.... }; \centerboxit{\box0}

% example of use: \boxtext{462pt}{This is the boxed text.}; 462pt is max length

% Picture macros and examples
%
% figures must be pasted from mcdraw
%
% Look in the 'Windows' menu for the pictures window
% It's like the Scrapbook -- cut and paste pictures
%

\def\picture #1 by #2 (#3){
  \vbox to #2{
    \hrule width #1 height 0pt depth 0pt
    \vfill
    \special{picture #3} % this is the low-level interface
    }
  }
% The first dimension of the picture macro is the width the second is depth

\def\scaledpicture #1 by #2 (#3 scaled #4){{
  \dimen0=#1 \dimen1=#2
  \divide\dimen0 by 1000 \multiply\dimen0 by #4
  \divide\dimen1 by 1000 \multiply\dimen1 by #4
  \picture \dimen0 by \dimen1 (#3 scaled #4)}
  }

%
% Note that you can also say, e.g.,
%  \special{postscript xxx yyy zzz}
% to include PostScript graphics in your documents
%
%Examples of use
%\def\stripes{\picture 2.29in by 1.75in (AWstripes)}
% By executing \stripes 
%\def\annie{\scaledpicture 102pt by 239pt (annie scaled 2000)}
%\def\finder{\picture 260pt by 165pt (screen0 scaled 500)}
%\def\icon{\picture 7in by 7in (icon)}
%Example of use
%\annie
%Example of centered picture \line{\hfil\annie\hfil}

%Figure w/  caption macro
\long\def\captfig#1#2#3#4#5{\vbox{\vskip1pc
\hbox to\hsize{\hfill{\picture #1 by #2 (#3)}\hfill}
\prevdepth=9pt \baselineskip=9pt
\vskip1pc
\hbox to\hsize{\hfill\vtop{\hsize=24pc\noindent{\eightbold Figure #4}
\hskip1em{\eightpoint#5}}\hfill}}}

%Examples of use of Figure macros
%\captfig{8.53pc}{19.9pc}{picturename}{5}{caption.}
%\captfig{2.23in}{2in}{picturename scaled 500}{16}{Caption.}
%The macro centers the picture.
%The first two numbers should be the true width
% and height after the picture has been scaled.
% So if the picture is scaled by 50% (500), the width and height in
% the macro should onw half of what they would be if the picture
% is not scaled (1000).
%
%
%
% Postcript macros

\def\illustration #1 by #2 (#3){
  \vskip#2\hskip#1\special{illustration #3} % this is the low-level interface
    }

\def\scaledillustration #1 by #2 (#3 scaled #4){{
  \dimen0=#1 \dimen1=#2
  \divide\dimen0 by 1000 \multiply\dimen0 by #4
  \divide\dimen1 by 1000 \multiply\dimen1 by #4
  \illustration \dimen0 by \dimen1 (#3 scaled #4)}
  }

% SHADEBOX.BSR MACROS
% Author: Leo@vaxc.cc.monash.edu.au
% Original Source:  Posted by Jimm Herreron <HERRERON@SMCVAX.BITNET>
% Modified from the file SHADEBOX.TEX on 9/30/93 by Becky Kaluza of Blue Sky
% Research to work with Textures 1.5 or later.

\newbox\graybox
\newdimen\xgrayspace
\newdimen\ygrayspace
%
% This macro can be used to typeset some text in a framed box with a
% shaded background. A set of examples can be found at the end of this
% file.
%
% This is a plain \TeX\ file modified for use on the Macintosh with Textures
% 1.5 or later.
%
% The characteristics of the shaded boxes are controlled by the following
% parameters
%
%   \xgrayspace = the space added before and after the text
%   \ygrayspace = the space above and below the text
%   \grayshade  = the gray colour 0 = black 1 = white
%   \linewidth  = the thickness of the border in points
%   \theradius  = the radius of the rounded corners in points
%   \thevskip   = extra \vskip added above and below the shaded box
%                 (applies only to \parashade)
%
%----------------------------------------------------------------------------
%
% The following \TeX code was adapted from previous work by
%
%            Je'ro^me Maillot, maillot@bora.inria.fr
%----------------------------------------------------------------------------
%
% Use the following for one or more words within a line.
%

\def\Textshade#1#2#3#4#5#6{%
    \xgrayspace=#4pt%
    \ygrayspace=#4pt%
    \def\grayshade{#3}%
    \def\linewidth{#5}%
    \def\theradius{#6}%
    \setbox\graybox=\hbox{\surroundboxa{#2}}%
    \hbox{%
    \hbox to 0pt{%
%!    \special{"gsave newpath 0 0 moveto                                %
    \PScommands
}%
    \box\graybox}}%
%
% Use the following for paragraphs.
%
\long%

\long%
\def\Parashade#1#2#3#4#5#6#7{%
    \xgrayspace=#4pt%
    \ygrayspace=#4pt%
    \def\grayshade{#3}%
    \def\linewidth{#5}%
    \def\theradius{#6}%
    \def\thevskip{#7pt}%
    \setbox\graybox=\hbox{\surroundboxb{#2}}%
    \vskip\thevskip%
    \hbox{%
    \hbox to 0pt{%
%!    \special{"gsave newpath 0 0 moveto                                %
    \PScommands
}%
     \box\graybox}%
     \vskip\thevskip%
}%
%----------------------------------------------------------------------------
%
% A pair of box macros. Each builds a slightly oversized box
% containing the text. The text is centred both in the vertical
% horizontal directions.
%
% Use the following for one or more words within a line.
%
\long\def\surroundboxa#1{\leavevmode\hbox{\vtop{%
\vbox{\kern\ygrayspace%
\hbox{\kern\xgrayspace#1%
      \kern\xgrayspace}}\kern\ygrayspace}}}
%
% Use the following for a paragraphs.
%
\long\def\surroundboxb#1{\leavevmode\hbox{\vtop{%
\vbox{\kern\ygrayspace%
\hbox{\kern\xgrayspace\vbox{\advance\hsize-2\xgrayspace#1}%
      \kern\xgrayspace}}\kern\ygrayspace}}}
%----------------------------------------------------------------------------
%
% Here are some simple PostScript routines.
%
% The TeX command \PScommands must be called before any of the
% shading routines can be used.
%
%!\long\def\PScommands{\special{! TeXDict begin
\long\def\PScommands{%
\special{rawpostscript
/sharpbox{%
           newpath
           xmin ymin moveto
           xmin ymax lineto
           xmax ymax lineto
           xmax ymin lineto
           xmin ymin lineto
           closepath 
          }bind def
}%
\special{rawpostscript
/sharpboxnb{%
           newpath
           xmin ymin moveto
           xmin ymax lineto
           xmax ymax lineto
           xmax ymin lineto
%           xmin ymin lineto
%           closepath 
          }bind def
}%
\special{rawpostscript
/sharpboxnt{%
           newpath
           xmin ymax moveto
           xmin ymin lineto
           xmax ymin lineto
           xmax ymax lineto
%           xmin ymin lineto
%           closepath 
          }bind def
}%
\special{rawpostscript
/roundbox{%
           newpath
           xmin radius add ymin moveto
           xmax ymin xmax ymax radius arcto
           xmax ymax xmin ymax radius arcto
           xmin ymax xmin ymin radius arcto
           xmin ymin xmax ymin radius arcto 16 {pop} repeat
           closepath
          }bind def
}%
\special{rawpostscript
/sharpcorners{%
               sharpbox gsave grayshade setgray fill grestore 
               linewidth setlinewidth stroke
              }bind def
}%
\special{rawpostscript
/sharpcornersnt{%
               sharpboxnt gsave grayshade setgray fill grestore 
               linewidth setlinewidth stroke
              }bind def
}%
\special{rawpostscript
/sharpcornersnb{%
               sharpboxnb gsave grayshade setgray fill grestore 
               linewidth setlinewidth stroke
              }bind def
}%
\special{rawpostscript
/roundcorners{%
               roundbox gsave grayshade setgray fill grestore 
               linewidth setlinewidth stroke
              }bind def
}%
\special{rawpostscript
/plainbox{%
           sharpbox grayshade setgray fill 
          }bind def
}%
% Here are the two new options
%
\special{rawpostscript
/roundnoframe{%
               roundbox grayshade setgray fill 
              }bind def
}%
\special{rawpostscript
/sharpnoframe{%
               sharpbox grayshade setgray fill 
              }bind def
}%
%!end}%
}%
%
% The \PScommands macro must be invoked before the shaded box macros.
%
%!\PScommands
% To use this, type \textshade{plainbox} or \textshade{roundbox} or
% \textshade{sharpbox}

%%%%% BOXES FOR TEXSHOP %%%%%

\def\boxit#1{\vbox{\hrule\hbox{\vrule\kern3pt
                                \vbox{\kern3pt#1\kern3pt}\kern3pt\vrule}\hrule}}
% example of use: \setbox0=\vbox{.... }; \boxit{\box0}

\def\boxitnb#1{\vbox{\hrule\hbox{\vrule\kern3pt
                                \vbox{\kern3pt#1\kern3pt}\kern3pt\vrule}}}

\def\boxitnt#1{\vbox{\hbox{\vrule\kern3pt
                                \vbox{\kern3pt#1\kern3pt}\kern3pt\vrule}\hrule}}

\def\boxitntnb#1{\vbox{\hbox{\vrule\kern3pt
                                \vbox{\kern3pt#1\kern3pt}\kern3pt\vrule}}}

% example of use: \boxtext{462pt}{This is the boxed text.}; 462pt is max length

% example of use: \boxtext{462pt}{This is the boxed text.}; 462pt is max length

% example of use: \boxtext{462pt}{This is the boxed text.}; 462pt is max length

% example of use: \boxtext{462pt}{This is the boxed text.}; 462pt is max length

%***************************************************
%         FONTS
%***************************************************

% ROMAN
%
%
%
%
%
%
%
%
\font\helbigbig=cmr10 scaled 2500%
\font\helbigb=cmbx10 scaled 1500%
\font\eightbold=cmbx8%

\def\tenf{\hel}%
\def\tenit{\heli}%
\def\ninef{\ninehel}%
\def\nineit{\nineheli}%
%
%

%  FONT FAMILIES

\font\tenrm=cmr10%
\font\teni=cmmi10%
\font\tensy=cmsy10%
\font\tenbf=cmbx10%
\font\tentt=cmtt10%
\font\tenit=cmti10%
\font\tensl=cmsl10%

\def\tenpoint{\def\rm{\fam0\tenrm}%
\textfont0=\tenrm%
\textfont1=\teni%
\textfont2=\tensy%
\textfont\itfam=\tenit%
\textfont\slfam=\tensl%
\textfont\ttfam=\tentt%
\textfont\bffam=\tenbf%
\scriptfont0=\sevenrm%
\scriptfont1=\seveni%
\scriptfont2=\sevensy%
%\scriptfont3=\tenex%
\scriptscriptfont0=\sixrm%
\scriptscriptfont1=\sixi%
\scriptscriptfont2=\sixsy%
%\scriptscriptfont3=\tenex%
\def\it{\fam\itfam\tenit}%
\def\tt{\fam\ttfam\tentt}%
\def\sl{\fam\slfam\tensl}%
\scriptfont\bffam=\sevenbf%
\scriptscriptfont\bffam=\sixbf%
\def\bf{\fam\bffam\tenbf}%
\normalbaselineskip=18pt%
\normalbaselines\rm}%

\font\ninerm=cmr9%
\font\ninebf=cmbx9%
\font\nineit=cmti9%
\font\ninesy=cmsy9%
\font\ninei=cmmi9%
\font\ninett=cmtt9%
\font\ninesl=cmsl9%

\def\ninepoint{\def\rm{\fam0\ninerm}%
\textfont0=\ninerm%
\textfont1=\ninei%
\textfont2=\ninesy%
\textfont\itfam=\nineit%
\textfont\slfam=\ninesl%
\textfont\ttfam=\ninett%
\textfont\bffam=\ninebf%
\scriptfont0=\sixrm%
\scriptfont1=\sixi%
\scriptfont2=\sixsy%
%\scriptfont3=\tenex%
\def\it{\fam\itfam\nineit}%
\def\tt{\fam\ttfam\ninett}%
\def\sl{\fam\slfam\ninesl}%
\scriptfont\bffam=\sixbf%
\scriptscriptfont\bffam=\fivebf%
\def\bf{\fam\bffam\ninebf}%
\normalbaselineskip=16pt%
\normalbaselines\rm}%

\font\eightrm=cmr8%
\font\eighti=cmmi8%
\font\eightsy=cmsy8%
\font\eightbf=cmbx8%
\font\eighttt=cmtt8%
\font\eightit=cmti8%
\font\eightsl=cmsl8%

\def\eightpoint{\def\rm{\fam0\eightrm}%
\textfont0=\eightrm%
\textfont1=\eighti%
\textfont2=\eightsy%
\textfont\itfam=\eightit%
\textfont\slfam=\eightsl%
\textfont\ttfam=\eighttt%
\textfont\bffam=\eightbf%
\scriptfont0=\sixrm%
\scriptfont1=\sixi%
\scriptfont2=\sixsy%
%\scriptfont3=\tenex%
\scriptscriptfont0=\fiverm%
\scriptscriptfont1=\fivei%
\scriptscriptfont2=\fivesy%
%\scriptscriptfont3=\tenex%
\def\it{\fam\itfam\eightit}%
\def\tt{\fam\ttfam\eighttt}%
\def\sl{\fam\slfam\eightsl}%
%\scriptfont\bffam=\sixbf%
\scriptscriptfont\bffam=\fivebf%
\def\bf{\fam\bffam\eightbf}%
\normalbaselineskip=14pt%
\normalbaselines\rm}%

\font\sevenrm=cmr7%
\font\seveni=cmmi7%
\font\sevensy=cmsy7%
\font\sevenbf=cmbx7%

\font\sixrm=cmr6%
\font\sixi=cmmi6%
\font\sixsy=cmsy6%
\font\sixbf=cmbx6%

\fontdimen13\tensy=2.6pt%
\fontdimen14\tensy=2.6pt%
\fontdimen15\tensy=2.6pt%
\fontdimen16\tensy=1.2pt%
\fontdimen17\tensy=1.2pt%
\fontdimen18\tensy=1.2pt%       

\def\tenf{\tenpoint}%
\def\ninef{\ninepoint}%
%

%%%%%%%%%%%% END OF MACROS %%%%%%%%%%%%

%%%%%%%%%% REDEFINITION OF BOX SPACING %%%%%%%%%%%%%%%%

\long\def\fig#1#2#3{\vbox{\vskip1pc\vskip#1
\prevdepth=12pt \baselineskip=12pt
\vskip1pc
\hbox to\hsize{\hfill\vtop{\hsize=30pc\noindent{\eightbf Figure #2\ }
{\eightpoint#3}}\hfill}}}

\def\show#1{}

\def\frac#1#2{{#1\over #2}}

\rightheadline{\botmark}

\pageno=1

%%%%%%%%%%%%%%%%%%%%%%%%%%%%%%%%%%%%%%%%%%%%

% GRAPHICS MACROS FOR TEX

%\input graphicx.tex

% DO NOT USE THE LATEX CODE

%\input miniltx
%
%\ifx\pdfoutput\undefined
 % \def\Gin@driver{dvips.def} % we are not running PDFTeX
%\else
%  \def\Gin@driver{pdftex.def} % we are running PDFTeX
%\fi

%\input graphicx.sty
%\resetatcatcode
%

% ALTERNATIVE SIMPLE METHOD

\input epsf

% Example of Use

%\epsfbox{PI_Minimax.eps}
%\centerline{\hskip0pc\epsfxsize = 4.5in \epsfbox{PI_Minimax.eps}}
%\centerline{\epsfscale=800 \epsfbox{PI_Minimax.eps}} NOTE: \epsfscale is questionable

%%%%%%%%%%%%%%%%%%%%%%%%%%%%%%%%%%%%%%%%%%%%

\pn {\bf December 2022}\hfill{\bf ASU Research Note; Ref: arXiv:2212.07998}
%\pn {\bf An updated version of a paper posted at arXiv preprint arXiv:22207.09588 on July 19, 2022.}
 %\hfill{\bf Report LIDS 2646.}
%\hfill {\bf Submitted for publication to JOTA}%
\bigskip \bigskip

\bigskip\bigskip\bigskip

\def\longpapertitle#1#2#3{{\bf \centerline{\helbigb
{#1}}}\medskip{\bf \centerline{\helbigb
{#2}}}\medskip{\centerline{
by}}\medskip{\bf \centerline{
{#3}}}\bigskip}

\longpapertitle{Rollout Algorithms and Approximate Dynamic Programming for}{Bayesian Optimization and Sequential Estimation}
{{Dimitri Bertsekas\ \footnote{\dag}{\ninepoint Fulton Professor of Computational Decision Making, School of Computing and Augmented Intelligence, Arizona State University, Tempe, AZ.}}}
\vskip1pc

\centerline{\bf Abstract}

We provide a unifying approximate dynamic programming framework that applies to a broad variety of problems involving sequential estimation. We consider first the construction of surrogate cost functions for the purposes of optimization, and we focus on the special case of Bayesian optimization, using the rollout algorithm and some of its variations. We then discuss the more general case of sequential estimation of a random vector using optimal measurement selection, and its application to problems of stochastic and adaptive control. We distinguish between adaptive control of deterministic and stochastic systems: the former are better suited for the use of rollout, while the latter are well suited for the use of rollout with certainty equivalence approximations. As an example of the deterministic case, we discuss sequential decoding problems, and a rollout algorithm for the approximate solution of the Wordle and Mastermind puzzles, recently developed in the paper [BBB22].

\vskip-1.5pc

\section{Bayesian Optimization}

 \pn In this paper, we consider a dynamic programming (DP) formulation of the problem of estimating an $m$-dimensional random vector $\theta=(\theta_1,\ldots,\theta_m)$, using optimal  sequential selection of observations, which are based on feedback from preceding observations. We discuss the use of reinforcement learning (RL) algorithms for solution, and rollout algorithms in particular. We will initially focus on the case of Bayesian optimization, particularly one involving a Gaussian process. The ideas extend more generally, as we will discuss in Section 4. The sequential estimation problem also arises in adaptive control formulations involving the optimal selection of both controls and observations to estimate the model parameters, as we explain in Section 5. As an example we discuss sequential decoding, whereby we search for a hidden code word by using a sequence of queries, in the spirit of the Wordle puzzle and the family of Mastermind games.
 
\xdef\figbograph{\figr}\figrnum\show{myfigure}

We will now discuss a DP framework for Bayesian optimization, where we aim to minimize a real-valued function $f$. The function is defined over a set of $m$ points, which we denote by $1,\ldots,m$. These $m$ points lie in some space, which we leave unspecified for the moment.\footnote{\dag}{\ninepoint We restrict the domain of definition of $f$ to be the finite set $\{1,\ldots,m\}$ in order to facilitate the implementation of the rollout algorithm to be discussed in what follows. However, in a more general formulation, the domain of $f$ can be an infinite set, such as a subset of a finite-dimensional Euclidean space.}  The value of $f$ at a point $u$  is denoted by $\theta_u$:
$$\theta_u=f(u),\qquad \hbox{for all }u=1,\ldots,m.$$
Thus the $m$-dimensional vector $\theta=(\theta_1,\ldots,\theta_m)$ belongs to $\re^m$ and represents the function $f$. In Bayesian optimization, we obtain sequentially noisy observations of values $f(u)=\theta_u$ at suitably selected points $u$. These values are used to estimate the vector $\theta$ (i.e., the function $f$), and to ultimately minimize (approximately) $f$ over the $m$ points $u=1,\ldots,m$. The essence of the problem is to  select points for observation  based on an exploration-exploitation tradeoff (exploring the potential of relatively unexplored candidate solutions and improving the estimate of promising candidate solutions). 

 \midinsert
\centerline{\hskip0pc\epsfxsize = 4.0in \epsfbox{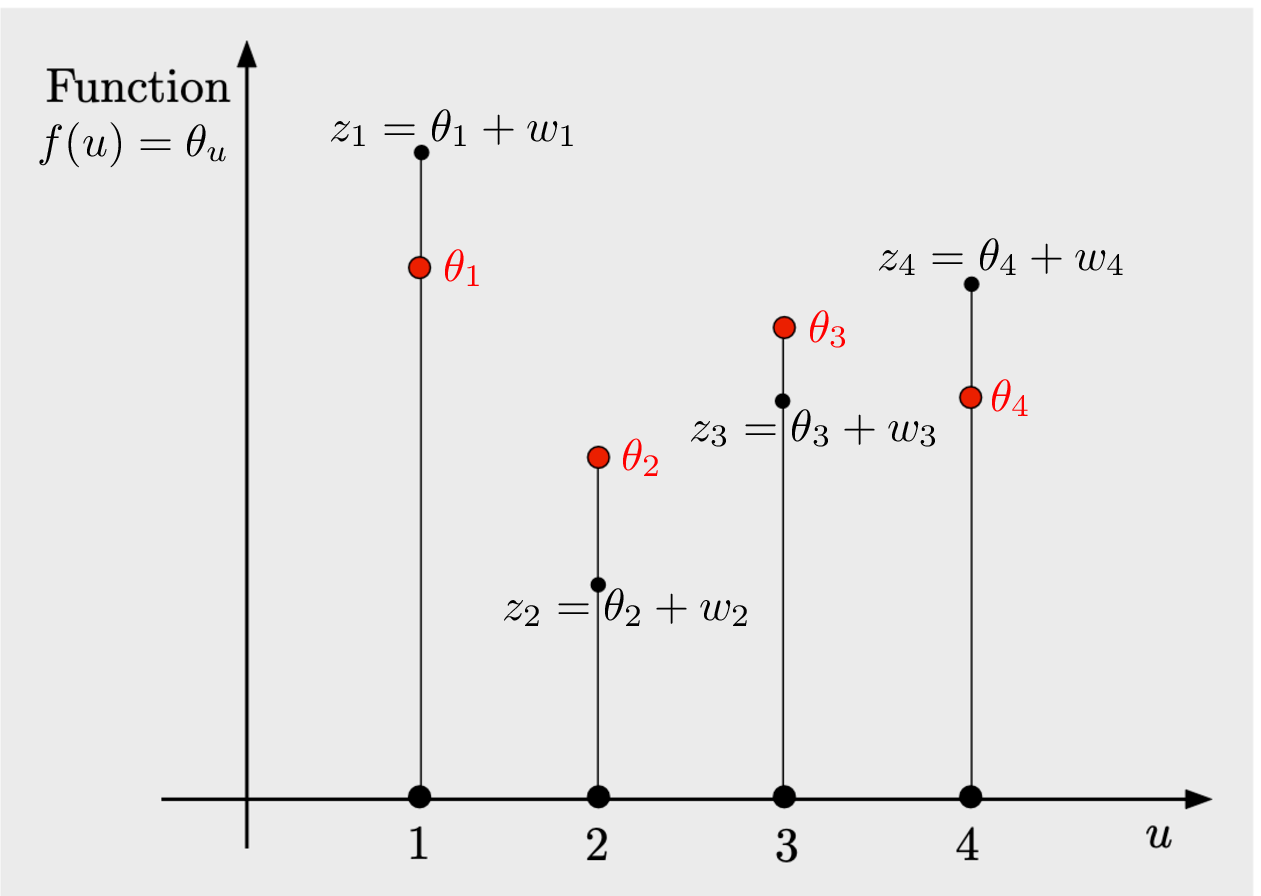}}
%\centerline{\includegraphics[width=4.0in]{bograph.eps}}
%\centerline{\hskip0pc\includegraphics[width=4.0in]{bograph}}
\fig{-1pc}{\figbograph}{Illustration of a function $f$ that we wish to estimate. The function is defined at the points $u=1,2,3,4$, and is represented by a vector $\theta=(\theta_1,\theta_2,\theta_3,\theta_4)\in\re^4$, in the sense that $f(u)=\theta_u$ for all $u$. The prior distribution of $\theta$ is given, and is used to construct the posterior distribution of $\theta$ given noisy observations $z_u=\theta_u+w_u$ at some of the points $u$.}
\endinsert

In particular, at each of $N$ successive times $k=1,\ldots,N$, we select a single point $u_k\in\{1,\ldots,m\}$, and observe the corresponding component $\theta_{u_k}$ of $\theta$ with some noise $w_{u_k}$, i.e.,
$$z_{u_k}=\theta_{u_k}+w_{u_k};\xdef\ukobservation{\lab}\eqnum\show{oneo}
$$
\smskip
\pn see Fig.\ \figbograph. We will view the observation points $u_1,\ldots,u_N$ as the optimization variables  (or controls/actions in a DP/RL context), and consider policies for selecting $u_k$ with knowledge of the preceding observations $z_{u_1},\ldots,z_{u_{k-1}}$ that have resulted from the  selections $u_1,\ldots,u_{k-1}$. We assume that the noise random variables $w_u$, $u\in \{1,\ldots,m\}$ are independent and that their distributions are given. Moreover, we assume that $\theta$ has a given a priori distribution on the space of $m$-dimensional vectors $\re^m$, which we denote by $b_0$. The posterior distribution of $\theta$, given any subset of observations 
$$\{z_{u_1},\ldots,z_{u_k}\},$$
is denoted by $b_k$. 

An important special case arises when $b_0$ and the distributions of $w_u$, $u\in \{1,\ldots,m\}$, are Gaussian. 
A key consequence of this assumption is that the posterior distribution $b_k$ is Gaussian, and can be calculated in closed form by using well-known formulas. More generally, $b_k$ evolves according to an equation of the form
$$b_{k+1}=B_k(b_k,u_{k+1},z_{u_{k+1}}),\qquad k=0,\ldots,N-1.\xdef\beliefpropone{\lab}\eqnum\show{oneo}$$
Thus given the set of observations up to time $k$, and the next choice $u_{k+1}$, resulting in an observation value $z_{u_{k+1}}$, the function $B_k$ gives the formula for updating $b_k$ to $b_{k+1}$, and may be viewed as a recursive estimator of $b_k$. In the Gaussian case, the function $B_k$ can be written in closed form, using standard formulas for Gaussian random vector estimation. In other cases where no closed form expression is possible, $B_k$ can be implemented  through simulation that computes (approximately) the new posterior $b_{k+1}$ using samples generated from the current posterior $b_k$.

At the end of the sequential estimation process, after the complete observation set 
$$\{z_{u_1},\ldots,z_{u_N}\}$$
has been obtained,  we have the posterior distribution $b_N$ of $\theta$, which we can use to compute a surrogate of $f$. As an example we may use as surrogate the posterior mean 
$\hat \theta=(\hat \theta_1,\ldots,\hat \theta_m)$, and declare as minimizer of $f$ over $u$ the point $u^*$ with minimum posterior mean:
$$u^*\in\arg\min\{\hat \theta_u\mid u=1,\ldots,m\}.$$

In the next section, we will focus on a DP formulation and a corresponding rollout algorithm for suboptimal sequential choice of observations, based on the results of the preceding observations. The DP approach draws its origin from long standing research on statistical design of sequential experiments, as well as related research on stochastic optimal control problems, where we aim to  find a policy that simultaneously optimizes over  observations and controls; see Section 5.

\vskip-1pc

\section{A Dynamic Programming Formulation}

\pn The sequential estimation problem as described in the preceding section, viewed as a DP problem, involves a state at time $k$, which is the posterior $b_k$, and a control/action at time $k$, which is the point index $u_{k+1}$ selected for observation. The transition equation according to which the state evolves, is 
$$b_{k+1}=B_k(b_k,u_{k+1},z_{u_{k+1}}),\qquad k=0,\ldots,N-1;$$
cf.\ Eq.\ \beliefpropone. To complete the DP formulation, we need to introduce a cost structure. To this end, we assume that observing $\theta_u$, as per Eq.\ \ukobservation, incurs a cost $c(u)$, and that there is a terminal cost $G(b_N)$ that depends of the final posterior distribution; as an example, the function $G$ may involve the mean and covariance of $b_N$.

The corresponding DP algorithm is \footnote{\dag}{\ninepoint An alternative DP formulation may be defined on the space of ``information vectors" $I_k$ at time $k$, given by $I_k=(z_{u_1},\ldots,z_{u_k},u_1,\ldots,u_k)$. This DP algorithm has the form
$$\skew5\hat J_k(I_k)=\min_{u_{k+1}\in\{1,\ldots,m\}}\Bigg[c(u_{k+1})+E_{z_{u_{k+1}}}\Big\{\skew5\hat J_{k+1}(I_k,z_{u_{k+1}},u_{k+1})\big)\ \big |\ I_k,u_{k+1}\Big\}\Bigg],\qquad k=0,\ldots,N-1,$$
and can similarly be used as the starting point for approximations.}
$$J_k^*(b_k)=\min_{u_{k+1}\in\{1,\ldots,m\}}\Bigg[c(u_{k+1})+E_{z_{u_{k+1}}}\Big\{J_{k+1}^*\big(B_k(b_k,u_{k+1},z_{u_{k+1}})\big)\ \big |\ b_k,u_{k+1}\Big\}\Bigg],\qquad k=0,\ldots,N-1,\xdef\dpequation{\lab}\eqnum\show{oneo}$$
and proceeds backwards from the terminal condition
$$J_N^*(b_N)=G(b_N).\eqnum\show{oneo}$$
Here the expected value in the right side of the DP equation \dpequation\ is taken with respect to the conditional distribution of $z_{u_{k+1}}$, given $b_k$ and the choice $u_{k+1}$. The observation cost $c(u)$ may be 0 or a constant for all $u$, and the terminal cost $G(b_N)$ may be a suitable measure of surrogate ``fidelity" that depends on the posterior mean and covariance of $\theta$ corresponding to $b_N$.

Generally, executing the DP algorithm \dpequation\ is practically infeasible, because the space of posterior distributions over $\rn$ is infinite-dimensional and very complicated. In the Gaussian case where the a priori distribution $b_0$ is Gaussian and the noise variables $w_u$ are Gaussian, the posterior $b_k$ is $m$-dimensional Gaussian, so it is characterized by its mean and covariance, and can be specified by a finite set of numbers. Despite this simplification, the  DP algorithm \dpequation\ is prohibitively time-consuming even under Gaussian assumptions, except for simple special cases. We consequently resort to DP/RL methods of approximation in value space, whereby the function $J_{k+1}^*$ in the right side of Eq.\ \dpequation\ is replaced by an approximation $\tl J_{k+1}$, as we will discuss in the next section. 

\vskip-1pc

\section{Approximation in Value Space and the Rollout Approach}

\pn The most popular Bayesian optimization methodology makes use of a myopic/greedy policy $\m_{k+1}$, which at each time $k$ and given $b_k$, selects a point $\hat u_{k+1}=\m_{k+1}(b_k)$ for the next observation, using some calculation involving an {\it acquisition function\/}. This function, 
denoted 
$A_k(b_k,u_{k+1})$, quantifies an ``expected benefit" for an observation at  $u_{k+1}$, given the current posterior $b_k$. The myopic policy selects the next point at which to observe, $\hat u_{k+1}$, by maximizing the acquisition function:
$$\hat u_{k+1}\in\arg\max_{u_{k+1}\in\{1,\ldots,m\}}A_k(b_k,u_{k+1}).\xdef\surbasepol{\lab}\eqnum\show{spconst}$$
Several ways to define suitable acquisition functions have been proposed, and an important issue is to be able to calculate economically its values $A_k(b_k,u_{k+1})$ for the purposes of the maximization in Eq.\ \surbasepol. Another important issue of course is to be able to calculate the posterior $b_k$ economically. There is a large literature on this methodology and its applications, particularly for the Gaussian case. We refer to the books by Rasmussen and Williams [RaW06], Powell and Ryzhov [PoR12], the highly cited papers by Saks et al.\ [SWM89],  and Queipo et al.\ [QHS05], the reviews  by Sasena [Sas02], Powell and Frazier [PoF08], Forrester and Keane [FoK09], Brochu, Cora, and De Freitas [BCD10], Ryzhov, Powell, and Frazier [RPF12], Ghavamzadeh, Mannor, Pineau, and Tamar [GMP15],  Shahriari et al.\ [SSW16], and Frazier [Fra18], and the references quoted there.

Approximation in value space is an alternative approach, which is based on the DP formulation of the preceding section. In particular, in this approach we approximate the DP algorithm \dpequation\ by replacing 
$J_{k+1}^*$ with an approximation $\tl J_{k+1}$ in the minimization of the right side. Thus we select the next observation at point $\tl u_{k+1}$ according to
$$\tl u_{k+1}\in\arg\min_{u_{k+1}\in\{1,\ldots,m\}}Q_k(b_k,u_{k+1}),\qquad k=0,\ldots,N-1,\xdef\dpapproxvspace{\lab}\eqnum\show{oneo}$$
where $Q_k(b_k,u_{k+1})$ is the Q-factor corresponding to the pair $(b_k,u_{k+1})$, given by
$$Q_k(b_k,u_{k+1})=c(u_{k+1})+E_{z_{u_{k+1}}}\Big\{\tl J_{k+1}\big(B_k(b_k,u_{k+1},z_{u_{k+1}})\big)\ \big |\ b_k,u_{k+1}\Big\},\qquad k=0,\ldots,N-1.\xdef\qfactor{\lab}\eqnum\show{oneo}$$
The expected value in the preceding equation is taken with respect to the conditional probability distribution of $z_{u_{k+1}}$ given $(b_k,u_{k+1})$, which can be computed using $b_k$ and the given distribution of the noise $w_{u_{k+1}}$. Thus if $b_k$ and $\tl J_{k+1}$ are available, we may use Monte Carlo simulation to determine the Q-factors $Q_k(b_k,u_{k+1})$ for all $u_{k+1}\in\{1,\ldots,m\}$, and select as next point for observation the one that corresponds to the minimal Q-factor [cf.\ Eq.\ \dpapproxvspace].

\xdef\figborollout{\figr}\figrnum\show{myfigure}

\subsection{Rollout Algorithms for Bayesian Optimization}

\pn A special case of approximation in value space is the rollout algorithm, whereby the function $J_{k+1}^*$ in the right side of the DP Eq.\ \dpequation\ is replaced by the cost function of some policy $\m_{k+1}(b_k)$, $k=0,\ldots,N-1$, called {\it base policy\/}. Thus, given a base policy  the rollout algorithm uses the cost function of this policy as the function $\tl J_{k+1}$ in the approximation in value space scheme \dpapproxvspace-\qfactor. The principal advantages of rollout are that it is well suited for the use of simulation, it overcomes the difficulties caused by the complexity of the state space, and with proper implementation it is typically supported by a cost improvement guarantee (the policy obtained by approximation in value space using some base policy performs better than the original base policy). The  values of $\tl J_{k+1}$ needed for the Q-factor calculations in Eq.\ \qfactor\ can be computed or approximated by simulation. Greedy/myopic policies based on an acquisition function [cf.\ Eq.\ \surbasepol] have been suggested as base policies in various rollout proposals.\footnote{\dag}{\ninepoint  The rollout algorithm for Bayesian optimization under Gaussian assumptions was first proposed by Lam, Wilcox, and Wolpert [LWW16]. This was further discussed by Jiang et al.\ [JJB20], [JCG20], Lee at al.\ [LEC20], Lee [Lee20], Yue and Kontar [YuK20], Lee et al.\ [LEP21], Paulson, Sorouifar, and Chakrabarty [PSC22], where it is also referred to as ``nonmyopic Bayesian optimization" or ``nonmyopic sequential experimental design." For related work, see  Gerlach, Hoffmann, and Charlish [GHC21]. These papers also discuss various approximations to the rollout approach, and generally report encouraging computational results.  
Extensive accounts of rollout algorithms and related RL subjects may be found in the author's DP book [Ber17], and RL books [Ber19], [Ber20a], where a cost improvement guarantee  is extensively discussed. Section 3.5 of the book [Ber20a] focuses on rollout algorithms for surrogate and Bayesian optimization.}

 \midinsert
\centerline{\hskip0pc\epsfxsize = 4.3in \epsfbox{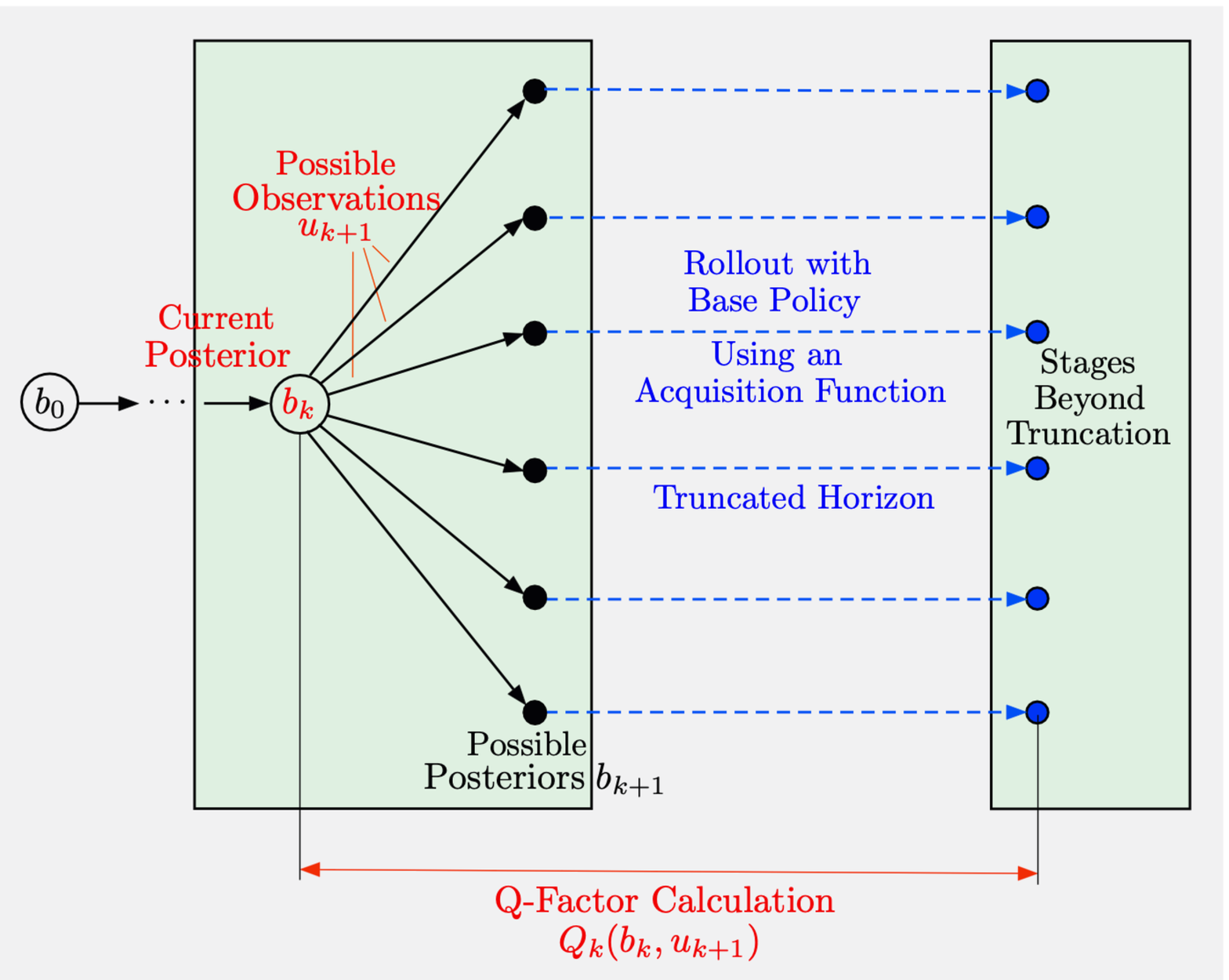}}
%\centerline{\includegraphics[width=4.6in]{borollout.eps}}
%\centerline{\hskip0pc\includegraphics[width=4.6in]{borollout}}
\fig{-1pc}{\figborollout}{Illustration of rollout at the current posterior $b_k$. For each $u_{k+1}\in\{1,\ldots,m\}$, we compute the Q-factor $Q_k(b_k,u_{k+1})$ by using Monte-Carlo simulation with samples from $w_{u_{k+1}}$  and a base heuristic that uses an acquisition function starting from each possible posterior $b_{k+1}$. The rollout may extend to the end of the horizon $N$, or it may be truncated after a few steps.}\endinsert

In particular, given $b_k$, the rollout algorithm computes for each $u_{k+1}\in\{1,\ldots,m\}$ a Q-factor value $Q_k(b_k,u_{k+1})$ by simulating the base policy for multiple time steps starting from all possible posteriors $b_{k+1}$ that can be generated from $(b_k,u_{k+1})$, and by accumulating the corresponding cost [including a terminal cost such as $G(b_N)$]; see Fig.\ \figborollout. It then selects the next point $\tl u_{k+1}$ for observation by using the Q-factor minimization of Eq.\ \dpapproxvspace. 

Note that the equation
$$b_{k+1}=B_k(b_k,u_{k+1},z_{u_{k+1}}),\qquad k=0,\ldots,N-1,$$
which governs the evolution of the posterior distribution, is stochastic because $z_{u_{k+1}}$ involves the stochastic noise $w_{u_{k+1}}$. Thus some Monte Carlo simulation is unavoidable in the calculation of the Q-factors $Q_k(b_k,u_{k+1})$. On the other hand, one may greatly reduce the Monte Carlo computational burden by employing a ``certainty equivalence" approximation, which at stage $k$,  treats only the noise $w_{u_{k+1}}$ as stochastic, and replaces the noise variables $w_{u_{k+2}}, w_{u_{k+3}}, \ldots$, after the first stage of the calculation, by deterministic quantities such as their means $\hat w_{u_{k+2}}, \hat w_{u_{k+3}}, \ldots$. This idea has been suggested for stochastic scheduling problems in the paper by Bertsekas and Castanon [BeC99], and in greater detail in the author's books [Ber20a] (Sections 2.2.4, 2.5.2, and Example 3.1.4) and [Ber22] (Sections 3.2 and 5.4) as an economical approximation of stochastic rollout, which does not degrade appreciably its performance.  The rationale is that even with this certainty equivalence approximation, the stochastic first step of the simulation ensures that rollout acts as a legitimate Newton step for solving the associated Bellman equation; we refer to [Ber20a] and [Ber22] for an extensive discussion of this issue. 

The simulation of the  Q-factor values may also involve other approximations, some of which have been suggested in various proposals on rollout-based Bayesian optimization. For example, if the number of possible observations $m$ is very large, we may compute and compare the Q-factors of only a subset. In particular, at a given time $k$, we may rank the observations by using an acquisition function, select a subset $U_{k+1}$ of most promising observations, compute their Q-factors $Q_k(b_k,u_{k+1})$, $u_{k+1}\in U_{k+1}$, and select the observation  whose Q-factor is minimal; this idea has been used in the case of the Wordle puzzle  [BBB22], which will be discussed later in Section 5.

\subsection{Multiagent Rollout for Bayesian Optimization}

\pn In some Bayesian optimization applications there arises the possibility of simultaneously performing multiple observations before receiving feedback about the corresponding observation outcomes. This occurs, among others, in two important contexts:

\nitem{(a)} In parallel computation settings, where multiple processors are used to perform simultaneously expensive evaluations of the function $f$ at multiple points $u$. These evaluations may involve some form of truncated simulation, so they yield evaluations of the form $z_u=\theta_u+w_u$, where $w_u$ is the simulation noise. 

\nitem{(b)} In distributed sensor systems, where a number of sensors provide in parallel relevant information about the random vector $\theta$ that we want to estimate; see e.g., the recent papers by Li, Krakow, and Gopalswamy [LKG21], [LKG22], which describe related multisensor estimation problems. 
\smskip

Of course in such  cases we may treat the entire set of simultaneous observations as a single observation within an enlarged Cartesian product space of observations, but there is a fundamental difficulty: the size of the observation space (and hence the number of Q-factors to be calculated by rollout at each time step) grows exponentially with the number of simultaneous observations. This in turn  greatly increases the computational requirements of the rollout algorithm.

To address this difficulty, we may employ the methodology of multiagent rollout, recently developed by the author in the papers [Ber20b], [Ber21], [BKB20], and the books [Ber20a], [Ber22].  This methodology addresses DP problems where the control at each stage consists of several distinct decisions, each one made by one of several agents. In multiagent rollout, the policy improvement is done one-agent-at-a-time in a given order, with (possibly partial) knowledge of the choices of the preceding agents in the order. As a result, the amount of computation for each policy improvement grows linearly with the number of agents, as opposed to exponentially for the standard all-agents-at-once method. At the same time the theoretical cost improvement property of the rollout algorithm can be shown to be preserved, while the empirical evidence suggests that great computational savings are achieved with hardly any performance degradation.  The book [Ber20a], and the papers [BKB20] and [Ber21] discuss the use of distributed schemes, whereby the agents do not requite full knowledge of the choices of the preceding agents, thereby increasing the potential for parallelism of the computations. 

\vskip-1.5pc

\section{Generalization to Sequential Estimation of Random Vectors}

\pn A more general problem that contains Bayesian optimization as a special case and admits a similar treatment, is to sequentially estimate an $m$-dimensional random vector $\theta=(\theta_1,\ldots,\theta_m)$ by using $N$ linear observations of $\theta$ of the form
$$z_u=a_u'\theta+w_u,\qquad u\in \{1,\ldots,n\},$$
where $n$ is some integer. Here $w_u$ are independent random variables with given probability distributions, the $m$-dimensional vectors $a_u$ are known, and $a_u'\theta$ denotes the inner product of $a_u$ and $\theta$. Similar to the case of Bayesian optimization, the problem simplifies if the given a priori distribution of $\theta$ is Gaussian, and  the random variables $w_u$ are independent and Gaussian.
Then, the posterior distribution of $\theta$, given any subset of observations, is Gaussian (thanks to the linearity of the observations), and can be calculated in closed form. 

Observations are generated sequentially at times $1,\ldots,N$, one at a time and with knowledge of the outcomes of the preceding observations, by choosing an index $u_k\in \{1,\ldots,n\}$ at time $k$, at a cost $c(u_k)$. Thus $u_k$ are the optimization variables, and affect both the quality of estimation of $\theta$ and the observation cost. The objective, roughly speaking, is to select $N$ observations to estimate $\theta$ in a way that minimizes an appropriate cost function; for example, one that penalizes some form of estimation error plus the cost of the observations. We can similarly formulate the corresponding optimization problem in terms of $N$-stage DP, and develop rollout algorithms for its approximate solution.

\vskip-1pc

\section{Adaptive Control of Systems with Unknown Parameters}

\pn In this section, we discuss the adaptive control of a dynamic system with unknown parameters. Our formulation involves simultaneous control of the system state and estimation of the unknown parameters. 
In such problems there may be multiple observation options, so there may be a tradeoff between selecting expensive/more informative observations and inexpensive/less informative observations. Moreover, the options for observation may be affected by the current state of the system, so possibilities may arise to steer the system towards states that are favorable for observation purposes. Bayesian optimization can be viewed as a special case where the decision just specifies the choice of observation at each step, while the system equation is suitably simplified, as will be discussed shortly. 

The idea of optimal observation selection in the context of stochastic optimal control was introduced in multiple works from the 60s. For some modern works that include the use of rollout, see   He and Chong [HeC06],  Hero et al.\ [HCC07], Chong, Kreucher, and Hero [CKH09], Jia [Jia10], Antunes and Heemels [AnH14], Beyme and Leung [BeL15], Khashooei, Antunes, and Heemels  [KAH15],   Gommans et al.\ [GTA17],  Elsherif, Chong, and Kim ECK19], Hoffmann et al.\ [HSC19], Molin,  Esen, and Johansson [MEJ19], Hoffmann et al.\ [HCR21], Lei et al.\ [LZY22], and the references given there.

The adaptive control formulation of this section  involves a state that consists of two components:
\nitem{(a)} A perfectly observed component $x_k$ that evolves over time according to a discrete-time system equation.

\nitem{(b)} An unknown parameter $\theta$,  which belongs to some space, and  is unobserved but stays constant. The perfectly observed component $x_k$ depends on $\theta$, and can be used to estimate $\theta$.
\smskip

\pn We view $\theta$ as a parameter in the system equation that governs the evolution of $x_k$.
 Thus we have a system of the form
$$x_{k+1}=f_k(x_k,\theta,u_k,w_k),\qquad k=0.\ldots,N-1,\xdef\system{\lab}\eqnum\show{oneo}$$
where $u_k$ is the control at time $k$, selected from a set $U_k(x_k)$, and $w_k$ is a random disturbance with given probability distribution that depends on $(x_k,\theta,u_k)$. The a priori probability distribution of $\theta$ is given and is updated based on the observed values of the state components $x_k$ and the applied controls $u_k$.\footnote{\dag}{\ninepoint  We do not address the complex measurability issues needed to make our problem definition mathematically rigorous. In the algorithms to be presented in this section, we will assume that $\theta$ takes values within a finite set.}  We also introduce a cost $g_k(x_k,\theta,u_k,w_k)$ for each stage $k=0,\ldots,N-1$, and a terminal cost $g_N(x_N)$, and view the problem as a Partially Observed Markovian Decision Problem (POMDP) with a special structure.  

Note that $x_k$ may involve components that play the role of observations. For example, $\theta$ may be a vector $(\theta_1,\ldots,\theta_m)$ in $\re^m$, $u_k$ may correspond to selection of one out of $m$ possible noisy observations of $\theta_u$, $u\in \{1,...,m\}$, and the system equation may have the form
$$x_{k+1} =\theta_{u_k} +w_{u_k};$$
cf.\ Eq.\ \ukobservation. This corresponds to the Bayesian optimization problem of Sections 1-3. Thus the adaptive
control formulation of the present section includes Bayesian optimization as a special case.

In principle $\theta$ may take values within a finite or an infinite set (such as $\re^m$). However, to facilitate the presentation of the subsequent DP analysis, and corresponding approximation in value space and rollout algorithms, we will  assume that $\theta$ can take one of $n$ known values $\theta^1,\ldots,\theta^n$ from within some unspecified space:
$$\theta\in\{\theta^1,\ldots,\theta^n\}.$$
Moreover, we assume that the information vector
$$I_k=(x_0,\ldots,x_k,u_0,\ldots,u_{k-1})$$
is available at time $k$, and is used to compute the posterior probabilities
$$b^i_{k}=P\{ \theta=\theta^i\mid I_k\},\qquad i=1,\ldots,n,$$
and the posterior distribution 
$$b_k=(b^1_{k},\ldots,b^n_{k}).$$
Together with the perfectly observed state $x_k$, the posterior distribution $b_k$ forms the pair $(x_k,b_k)$, which is commonly called the {\it belief state} of the POMDP at time $k$. 

Note that according to the classical methodology of POMDP (see e.g., [Ber17a], Chapter 4), the belief component $b_{k+1}$ is determined by the belief state $(x_k,b_k)$, the control $u_k$, and the observation obtained at time $k+1$, i.e., $x_{k+1}$. Thus $b_k$ can be updated according to an equation of the form
$$b_{k+1}=B_k(x_k,b_k,u_k,x_{k+1}),$$
where $B_k$ is an appropriate function, 
which can be viewed as a recursive estimator of $\theta$. There are several approaches to implement this estimator (perhaps with some approximation error), including closed formed expressions that involve the use of Bayes' rule, as well as the simulation-based method of particle filtering.

\subsection{The Exact DP Algorithm - Approximation in Value Space}

\pn We will now describe an exact DP algorithm that operates in the space of information vectors $I_k$. To describe this algorithm, let us denote by $J^*_k(I_k)$ the optimal cost starting at information vector $I_k$ at time $k$. Using the equation
$$I_{k+1}=(I_k,x_{k+1},u_k)=\big(I_k,f_k(x_k,\theta,u_k,w_k),u_k\big),$$
the algorithm takes the form
$$\eqalign{J^*_k(I_k)=&\min_{u_k\in U_k(x_k)}E_{\theta,w_k}\Big\{g_k(x_k,\theta,u_k,w_k)+\cr
&\ \ \ \ \ \ \ \ \ \ \ \ \ \ \ \ \ \ \ \ \ \ \ \ \ \ \ \ \ \ \ \ J^*_{k+1}\big(I_k,f_k(x_k,\theta,u_k,w_k),u_k\big)\mid I_k,u_k\Big\},\cr}\xdef\dpalgorithmorig{\lab}\eqnum\show{oneo}$$
for $k=0,\ldots,N-1$, with $J^*_N(I_N)=g_N(x_N)$; see e.g., the DP textbook [Ber17a], Section 4.1. 

By using the law of iterated expectations, 
$$E_{\theta,w_k}\{\cdot\mid I_k,u_k\}=E_\theta\big\{E_{w_k}\{\cdot\mid I_k,\theta,u_k\}\mid I_k,u_k\big\},$$
we can rewrite this DP algorithm as
$$\eqalign{J^*_k(I_k)=&\min_{u_k\in U_k(x_k)}\sum_{i=1}^n b^i_{k}E_{w_k}\Big\{g_k(x_k,\theta^i,u_k,w_k)+\cr
&\ \ \ \ \ \ \ \ \ \ \ \ \ \ \ \ \ \ \ \ \ \ \ \ \ \ \ \ \ \ \ \ \ \ \ \ J^*_{k+1}\big(I_k,f_k(x_k,\theta^i,u_k,w_k),u_k\big)\mid I_k,\theta^i,u_k\Big\}.\cr}\xdef\dpalgorithm{\lab}\eqnum\show{oneo}$$
The summation over $i$ above represents the expected value over $\theta$ conditioned on $I_k$ and $u_k$. 

The algorithm \dpalgorithm\ is typically very hard to implement, because of the dependence of $J^*_{k+1}$ on the entire information vector $I_{k+1}$, which expands in size according to
$$I_{k+1}=(I_k,x_{k+1},u_k).$$
To address this implementation difficulty, we may use approximation in value space, based on replacing $J^*_{k+1}$ in the DP algorithm \dpalgorithmorig\  with some function that can be obtained (either off-line or on-line) with a tractable computation. 

One approximation possibility is based on the use of the optimal cost function corresponding to each parameter value $\theta^i$,
$$\hat J_{k+1}^i(x_{k+1}),\qquad i=1,\ldots,n.\xdef\optcosts{\lab}\eqnum\show{oneo}$$
Here, $\hat J_{k+1}^i(x_{k+1})$ is the optimal cost that would be obtained starting from state $x_{k+1}$ under the assumption that $\theta$ is known to be equal to $\theta^i$; this corresponds to a perfect state information problem. Then, given $x_k$ and the posterior distribution $b_k$ (which is detemined from $I_k$), an approximation in value space scheme with one-step lookahead minimization applies at time $k$ a control 
$$\eqalign{\tl u_k&\in\arg\min_{u_k\in U_k(x_k)}\sum_{i=1}^n b^i_{k}E_{w_k}\Big\{g_k(x_k,\theta^i,u_k,w_k)+\cr
&\ \ \ \ \ \ \ \ \ \ \ \ \ \ \ \ \ \ \ \ \ \ \ \ \ \ \ \ \ \ \ \ \ \ \ \ \ \ \ \ \ \ \ \ \ \ \ \hat J_{k+1}^i\big(f_k(x_k,\theta^i,u_k,w_k)\big)\mid x_k,\theta^i,u_k\Big\}.\cr}\xdef\lookscheme{\lab}\eqnum\show{oneo}$$
Thus, instead of the optimal control, which minimizes the optimal Q-factor of $(I_k,u_k)$ appearing in the right side of Eq.\ \dpalgorithmorig, we apply control $\tl u_k$ that minimizes the expected value over $\theta$ of the optimal Q-factors that correspond to fixed and known values of $\theta$.

A simpler version of this approach is to use the same function $\hat J_{k+1}^i$ for every $i$. However, the dependence on $i$ may be useful in some contexts where differences in the value of $i$ may have a radical effect on the qualitative character of the system equation.

Generally, the optimal costs $\hat J_{k+1}^i(x_{k+1})$ that correspond to the different parameter values $\theta^i$ [cf.\ Eq.\ \optcosts] may be hard to compute, despite their perfect state information structure.\footnote{\dag}{\ninepoint In favorable special cases, such as linear quadratic problems, the optimal costs $\skew6\hat J_{k+1}^i(x_{k+1})$ may be easily calculated in closed form. Still, however, even in such cases the calculation of the belief probabilities $b^i_{k}$ may not be simple, and may require the use of a system identification algorithm.} An alternative possibility is to use off-line trained feature-based or neural network-based approximations to $\hat J_{k+1}^i(x_{k+1})$. 

In an infinite horizon variant of the problem, it is reasonable to expect that the estimate of the parameter $\theta$ improves over time, and that with a suitable estimation scheme, it converges asymptotically to the correct value of $\theta$, call it $\theta^*$, i.e., 
$$ \lim_{k\to\infty} b^i_{k}=\cases{1&if $\theta^i=\theta^*$,\cr
0&if $\theta^i\ne\theta^*$.\cr}$$
Then it can be seen that the generated one-step lookahead controls $\tl u_k$ are asymptotically obtained from the Bellman equation that corresponds to the correct parameter $\theta^*$, and are typically optimal in some asymptotic sense. Schemes of this type have been discussed in the adaptive control literature since the 70s; see e.g., Mandl [Man74], Doshi and Shreve [DoS80], Kumar and Lin [KuL82], Kumar [Kum85]. Moreover, some of the pitfalls of performing parameter identification while simultaneously applying adaptive control have been described by Borkar and Varaiya [BoV79], and by Kumar [Kum83]; see [Ber17], Section 6.8 for a related discussion.

\subsection{Rollout}

\pn Another approximation possibility is to use the costs of given policies $\p^i$ in place of the optimal costs $\hat J_{k+1}^i(x_{k+1})$ in the approximation in value space scheme of Eq.\ \lookscheme. In this case the one-step lookahead scheme \lookscheme\ takes the form 
$$\eqalign{\tl u_k&\in\arg\min_{u_k\in U_k(x_k)}\sum_{i=1}^n b^i_{k}E_{w_k}\Big\{g_k(x_k,\theta^i,u_k,w_k)+\cr
&\ \ \ \ \ \ \ \ \ \ \ \ \ \ \ \ \ \ \ \ \ \ \ \ \ \ \ \ \ \ \ \ \ \ \ \ \ \ \ \ \ \ \ \ \ \ \ \hat J_{k+1,\p^i}^i\big(f_k(x_k,\theta^i,u_k,w_k)\big)\mid x_k,\theta^i,u_k\Big\},\cr}\xdef\policylookscheme{\lab}\eqnum\show{oneo}$$
and has the character of a rollout algorithm, with $\p^i=\{\m_0^i,\ldots,\m_{N-1}^i\}$, $i=1,\ldots,n$, being known base policies, with components $\m_k^i$ that depend only on $x_k$. Here, the term
$$\hat J_{k+1,\p^i}^i\big(f_k(x_k,\theta^i,u_k,w_k)\big)$$
in Eq.\ \policylookscheme\ is the cost of the base policy $\p^i$, calculated starting from the next state 
$$x_{k+1}=f_k(x_k,\theta^i,u_k,w_k),$$
under the assumption that $\theta$ will stay fixed at the value $\theta=\theta^i$ until the end of the horizon. These costs must be computed separately for each of the $n$ values $\theta^i$, and averaged using the belief distribution $b_k$ to form the Q-factors that are minimized in Eq.\ \policylookscheme\ to obtain the rollout control $\tl u_k$.

\subsection{The Case of a Deterministic System}

\pn Let us now consider the case where the system \system\ is deterministic of the form
$$x_{k+1}=f_k(x_k,\theta,u_k).\xdef\detsystem{\lab}\eqnum\show{oneo}$$
Then, while the problem still has a stochastic character due to the uncertainty about the value of $\theta$, the DP algorithm \dpalgorithm\ and its approximation in value space counterparts are greatly simplified because there is no expectation over $w_k$ to contend with. Indeed, given a state $x_k$, a parameter value $\theta^i$, and a control $u_k$, the on-line computation of the control of the rollout-like algorithm \policylookscheme,  takes the form
$$\tl u_k\in\arg\min_{u_k\in U_k(x_k)}\sum_{i=1}^n b^i_{k}\Big(g_k(x_k,\theta^i,u_k)+ \hat J_{k+1,\p^i}^i\big(f_k(x_k,\theta^i,u_k)\big)\Big).\xdef\policylookdetscheme{\lab}\eqnum\show{oneo}$$
Note that the term  
$$\hat J_{k+1,\p^i}^i\big(f_k(x_k,\theta^i,u_k)\big)$$ 
in the above equation must be computed for every pair $(u_k,\theta^i)$, with $u_k\in U_k(x_k)$, $i=1,\ldots,n$. However, its computation does not involve Monte Carlo simulation, and can be performed with a deterministic propagation from the state $x_{k+1}$ of Eq.\ \detsystem\ up to the end of the horizon, using the base policy $\p^i$, while assuming that $\theta$ is fixed at the value $\theta^i$. 

\xdef\figpomdprollout{\figr}\figrnum\show{myfigure}

\topinsert
\centerline{\hskip0pc\epsfxsize = 4.4in \epsfbox{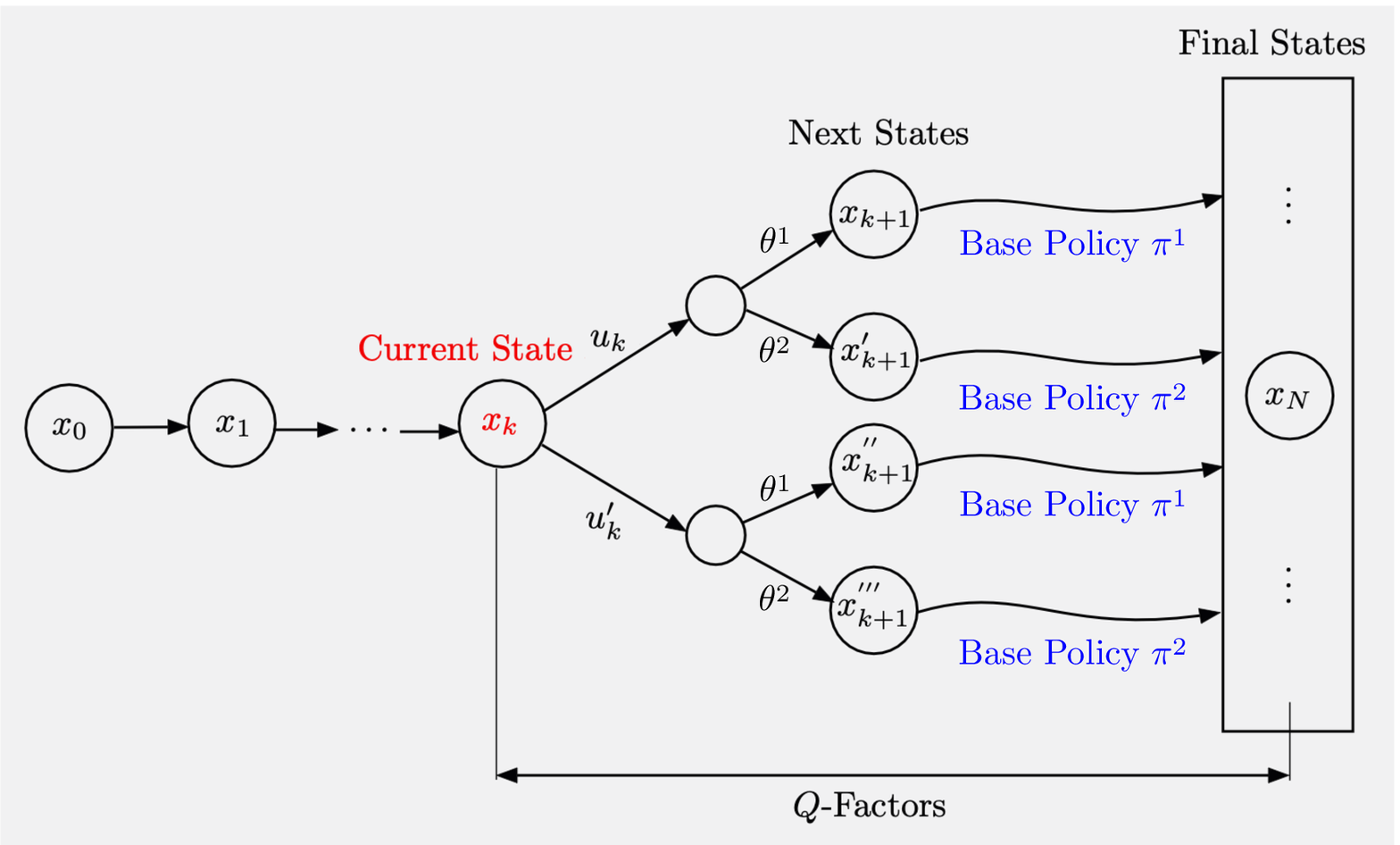}}
%\centerline{\includegraphics[width=4.5in]{pomdprollout_BO.eps}}
%\centerline{\hskip0pc\includegraphics[width=4.5in]{pomdprollout_BO}}
\vskip0pt
\fig{-1pc}{\figpomdprollout}{Schematic illustration of adaptive control by rollout for deterministic systems. The Q-factor $Q_k(u_k,\theta^i)$ is computed for every pair $(u_k,\theta^i)$ by using the base policy $\p^i$ and deterministic propagation of the system equation starting from the next state
$$x_{k+1}=f_k(x_k,\theta^i,u_k).$$
The Q-factor $Q_k(u_k,\theta^i)$  are then averaged over $\theta^i$, using the current belief distribution $b_k$, and the control applied is the one that minimizes the averaged Q-factor 
$$\sum_{i=1}^n b^i_{k}Q_k(u_k,\theta^i)$$
 over $u_k\in U_k(x_k)$.}
\endinsert

The term 
$$Q_k(u_k,\theta^i)=g_k(x_k,\theta^i,u_k)+ \hat J_{k+1,\p^i}^i\big(f_k(x_k,\theta^i,u_k)\big)$$
appearing on the right side of Eq.\ \policylookdetscheme\ can bet viewed as a Q-factor. Its expected value,
$$\hat Q_k(u_k)=\sum_{i=1}^n b^i_{k}Q_k(u_k,\theta^i),$$
must be calculated for every $u_k\in U_k(x_k)$, and the computation of the rollout control $\tl u_k$ is then obtained from the minimization
$$\tl u_k\in\arg\min_{u_k\in U_k(x_k)}\hat Q_k(u_k);$$
cf.\ Eq.\ \policylookdetscheme. This computation is illustrated in Fig.\ \figpomdprollout. 

The case of a deterministic system is particularly interesting because we can typically expect that the true parameter $\theta^*$ is identified in a finite number of stages, since at each stage $k$, we are receiving a noiseless measurement relating to $\theta$, namely the state $x_k$. Once this happens, the problem becomes one of perfect state information.

An intuitive example that demonstrates the deterministic system case is the popular Worlde puzzle.

\xdef\examplewordle{\exampl}\examplnum\show{examplo}

\beginexample{\examplewordle\ (The Wordle Puzzle)}Wordle is a decoding puzzle whereby we try to guess a mystery word $\theta^*$ out of a known finite collection of 5-letter words. This is done with sequential guesses each of which provides additional information on the correct word $\theta^*$, by using certain given rules to shrink the current mystery list (the smallest list that contains $\theta^*$, based on the currently available information). The objective is to minimize the number of guesses to find $\theta^*$. 

The rules for shrinking the mystery list relate to the common letters between the word guesses and the mystery word $\theta^*$, and they will not be described here (there is a large literature regarding Wordle). Moreover, $\theta^*$ is assumed to be chosen from the initial collection of 5-letter words according to a uniform distribution. Under this assumption, it can be shown that the belief distribution $b_k$ at stage $k$ continues to be uniform over the current mystery list. As a result, we may use as state $x_k$ the mystery list at stage $k$, which evolves deterministically according to an equation of the form \detsystem, where $u_k$ is the guess word at stage $k$. There are several base policies to use in the rollout-like algorithm \policylookdetscheme, which are described in the paper by Bhambri, Bhattacharjee, and Bertsekas [BBB22], together with computational results showing that the corresponding rollout algorithm  \policylookdetscheme\ performs remarkably close to optimal (within less than $0.5\%$ of the optimal average number of guesses, as calculated by Selby [Sel22]). 

We also note that the rollout approach applies to several variations of the Wordle puzzle for which an optimal solution is intractable. Such variations may include for example a larger length $\ell>5$ of mystery words, and/or a known nonuniform distribution over the initial collection of $\ell$-letter words; see [BBB22]. The rollout approach also applies to the popular family of Mastermind games that centers around finding a hidden sequence of objects (e.g., letters or colors) using partial observations, and shares the sequential decoding structure of the Wordle puzzzle.
\endexample

Finally, let us note that an illustration similar to the one of  Fig.\ \figpomdprollout\ applies for the case of the stochastic system of Eq.\ \system\ [cf.\ the rollout scheme \policylookscheme]. In this case, a Q-factor 
$$Q_k(u_k,\theta^i,w_k)=g_k(x_k,\theta^i,u_k,w_k)+\hat J_{k+1,\p^i}^i\big(f_k(x_k,\theta^i,u_k,w_k)\big)$$
must be calculated for every triplet $(u_k,\theta^i,w_k)$, using the base policy $\p^i$. The rollout  control  $\tl u_k$ is obtained by minimizing the expected value of this Q-factor [averaged using the distribution of $(\theta,w_k)$]; cf.\ Eq.\ \policylookscheme. However, for a stochastic system, in contrast to the deterministic case,  the calculation of the Q-factors 
$$Q_k(u_k,\theta^i,w_k)$$
 requires Monte Carlo simulation, to take into account the current and future noise terms $w_{k},\ldots,w_{N-1}$. The Monte Carlo computation cost may be mitigated by using a certainty equivalence approximation for the future noise terms
$w_{k+1},\ldots,w_{N-1}$, as noted in Section 3.1, albeit with additional potential loss of optimality.

\vskip-1.5pc

\section{Concluding Remarks}

\pn We introduced a unified framework for construction of surrogate cost functions by Bayesian optimization,  and for more general sequential estimation and control problems. A common attribute of these problems is the aim to estimate some hidden variables through the use of targeted sequential observations. We have focused on the rollout algorithm methodology, which is simple and well suited for this framework. Its principal characteristic is that it yields high quality approximate solutions, with  improved performance over commonly used heuristics, which can be used as base policies. Moreover, the rollout algorithm can be economically extended to multiagent problems, involving simultaneous observations, where the classical Bayesian optimization methods may be impractical.

While the rollout algorithm generally requires a substantial amount of Monte Carlo simulation, it admits parallelization and approximations based on truncation of its horizon and the use of certainty equivalence beyond the first step of the simulation. These approximations speed up the computations and enlarge the range of the practical applications of the rollout approach. 

\vskip-1.5pc

\section{References}
\vskip-0.5pc

\ref[AnH14] Antunes, D., and Heemels, W.\ P.\ M.\ H., 2014.\ ``Rollout Event-Triggered Control: Beyond Periodic Control Performance," IEEE Trans.\ on Automatic Control, Vol.\ 59, pp.\ 3296-3311.

\ref[BBB22] Bhambri, S., Bhattacharjee, A., and Bertsekas, D.\ P., 2022.\ ``Reinforcement Learning Methods for Wordle: A POMDP/Adaptive Control Approach," arXiv preprint arXiv:2211.10298.

\ref[BCD10] Brochu, E., Cora, V.\ M., and De Freitas, N., 2010.\ ``A Tutorial on Bayesian Optimization of Expensive Cost Functions, with Application to Active User Modeling and Hierarchical Reinforcement Learning," arXiv preprint arXiv:1012.2599.

\ref[BKB20] Bhattacharya, S., Kailas, S., Badyal, S., Gil, S., Bertsekas, D., 2020.\``Multiagent Rollout and Policy Iteration for POMDP with Application to   Multi-Robot Repair Problems," Proc. CORL.

\ref[BeC99] Bertsekas, D.\ P., and  Casta\~non, D.\ A., 1999.\ ``Rollout Algorithms for
Stochastic Scheduling Problems," Heuristics, Vol.\ 5, pp.\ 89-108. 

\ref[[BeL15] Beyme, S., and Leung, C., 2015.\ ``Rollout Algorithms for Wireless Sensor Network-Assisted Target Search," IEEE Sensors Journal, Vol.\ 15, pp.\ 3835-3845.

\ref[Ber17] Bertsekas, D.\ P., 2017.\ Dynamic Programming and Optimal Control, Vol.\ I, 4th Ed., Athena Scientific, Belmont, MA.

\ref[Ber19] Bertsekas, D.\ P., 2019.\ Reinforcement Learning and Optimal Control, Athena Scientific, Belmont, MA.

\ref[Ber20a] Bertsekas, D.\ P., 2020.\ Rollout, Policy Iteration, and Distributed Reinforcement Learning, Athena Scientific, Belmont, MA.

\ref[Ber20b] Bertsekas, D.\ P., 2020.\ ``Multiagent Value Iteration Algorithms in Dynamic Programming and Reinforcement Learning," Results in Control and Optimization J., Vol.\ 1.

\ref[Ber21] Bertsekas, D.\ P., 2021.\ ``Multiagent Reinforcement Learning: Rollout and Policy Iteration," IEEE/CAA Journal of Automatica Sinica, Vol. 8, pp.\ 249-271.

\ref[Ber22] Bertsekas, D.\ P., 2022.\ Lessons from AlphaZero for Optimal, Model Predictive, and Adaptive Control, Athena Scientific, Belmont, MA.

\ref [BoV79] Borkar, V., and Varaiya, P.\ P., 1979.\  ``Adaptive Control of
Markov Chains, I:  Finite Parameter Set," IEEE Trans.\ Automatic Control, Vol.\
AC-24, pp.\ 953-958.

\ref[CKH09] Chong, E.K., Kreucher, C.\ M. and Hero, A.\ O., 2009.\ ``Partially Observable Markov Decision Process Approximations for Adaptive Sensing, Discrete Event Dynamic Systems, Vol.\ 19, pp.\ 377-422.

\ref [DoS80] Doshi, B., and Shreve, S., 1980.\  ``Strong Consistency of a Modified
Maximum Likelihood Estimator for Controlled Markov Chains," J.\ of Applied
Probability, Vol.\ 17, pp.\ 726-734.

\ref[ECK19] Elsherif, F., Chong, E.\ K. and Kim, J.\ H., 2019.\ ``Energy-Efficient Base Station Control Framework for 5G Cellular Networks Based on Markov Decision Process," IEEE Transactions on Vehicular Technology, Vol.\ 68, pp.\ 9267-9279.

\ref[FoK09] Forrester, A.\ I., and Keane, A.\ J., 2009.\ ``Recent Advances in Surrogate-Based Optimization.\ Progress in Aerospace Sciences," Vol.\ 45, pp.\ 50-79.

\ref[Fra18] Frazier, P.\ I., 2018.\ ``A Tutorial on Bayesian Optimization," arXiv preprint arXiv:1807.02811.

\ref[GHC21] Gerlach, T., Hoffmann, F., and Charlish, A., 2021.\ ``Policy Rollout Action Selection with Knowledge Gradient for Sensor Path Planning," 2021 IEEE 24th International Conference on Information Fusion, pp. 1-8.

\ref[GMP15] Ghavamzadeh, M., Mannor, S., Pineau, J., and Tamar, A., 2015.\ ``Bayesian Reinforcement Learning: A Survey," Foundations and Trends in Machine Learning, Vol.\ 8, pp.\ 359-483.

\ref[GTA17] Gommans, T.\ M.\ P., Theunisse, T.\ A.\ F., Antunes, D.\ J., and Heemels, W.\ P.\ M.\ H., 2017.\ ``Resource-Aware MPC for Constrained Linear Systems: Two Rollout Approaches," Journal of Process Control, Vol.\ 51, pp.\ 68-83.

\ref[HCC07] Hero, A.\ O., Casta\~non, D., Cochran, D., and Kastella, K., eds., 2007.\ Foundations and Applications of Sensor Management, Springer Science.

\ref[HCR21] Hoffmann, F., Charlish, A., Ritchie, M., and Griffiths, H., 2021.\ ``Policy Rollout Action Selection in Continuous Domains for Sensor Path Planning," IEEE Transactions on Aerospace and Electronic Systems.

\ref[HSC19] Hoffmann, F., Schily, H., Charlish, A., Ritchie, M., and Griffiths, H., 2019.\ ``A Rollout Based Path Planner for Emitter Localization," 22th International Conference on Information Fusion.

\ref[HeC06] He, Y., and Chong, E.\ K., 2006.\ ``Sensor Scheduling for Target Tracking: A Monte Carlo Sampling Approach," Digital Signal Processing, Vol.\ 16, pp.\ 533-545.

\ref[JCG20] Jiang, S., Chai, H., Gonzalez, J., and Garnett, R., 2020.\ ``BINOCULARS for Efficient, Nonmyopic Sequential Experimental Design," In Proc.\ Intern.\ Conference on Machine Learning, pp.\ 4794-4803.

\ref[JJB20] Jiang, S., Jiang, D.\ R., Balandat, M., Karrer, B., Gardner, J.\ R., and Garnett, R., 2020.\ ``Efficient Nonmyopic Bayesian Optimization via One-Shot Multi-Step Trees," arXiv preprint arXiv:2006.15779.

\ref[Jia10] Jia, Q.\ S., 2010.\ ``A Rollout Method for Finite-Stage Event-Based Decision Processes," IFAC Proceedings Vol.\ 43(12), pp.\ 247-252.

\ref[KAH15] Khashooei, B.\ A., Antunes, D.\ J., and Heemels, W.\ P.\ M.\ H., 2015.\ ``Rollout Strategies for Output-Based Event-Triggered Control," In Proc.\ 2015 European Control Conference, pp.\ 2168-2173.

\ref [KuL82] Kumar, P.\ R., and Lin, W., 1982.\  ``Optimal Adaptive Controllers for
Unknown Markov Chains," IEEE Trans.\ Automatic Control, Vol.\ AC-27, pp.\ 765-774.

\ref [Kum83] Kumar, P.\ R., 1983.\  ``Optimal Adaptive
Control of Linear-Quadratic-Gaussian Systems," SIAM J.\ on Control and Optimization,
Vol.\ 21, pp.\ 163-178.

\ref [Kum85] Kumar, P.\ R., 1985.\  ``A Survey of Some Results in Stochastic Adaptive
Control," SIAM J.\ on Control and Optimization, Vol.\ 23, pp.\ 329-380.

\ref[LEC20] Lee, E.\ H., Eriksson, D., Cheng, B., McCourt, M., and Bindel, D., 2020.\ ``Efficient Rollout Strategies for Bayesian Optimization," arXiv preprint arXiv:2002.10539.

\ref[LEP21] Lee, E.\ H., Eriksson, D., Perrone, V., and Seeger, M., 2021.\ ``A Nonmyopic Approach to Cost-Constrained Bayesian Optimization," In Uncertainty in Artificial Intelligence Proceedings, pp.\ 568-577.

\ref[LKG21] Li, T., Krakow, L.\ W., and Gopalswamy, S., 2021.\ ``Optimizing Consensus-Based Multi-Target Tracking with Multiagent Rollout Control Policies," In 2021 IEEE Conference on Control Technology and Applications, pp.\ 131-137.

\ref[LKG22] Li, T., Krakow, L.\ W., and Gopalswamy, S., 2022.\ ``SMA-NBO: A Sequential Multi-Agent Planning with Nominal Belief-State Optimization in Target Tracking," arXiv preprint arXiv:2203.01507.

\ref[LWW16] Lam, R., Willcox, K., and Wolpert, D.\ H., 2016.\ ``Bayesian Optimization with a Finite Budget: An Approximate Dynamic Programming Approach," In Advances in Neural Information Processing Systems, pp.\ 883-891.

\ref[LZY22] Lei, W., Zhang, D., Ye, Y., and Lu, C., 2022.\ ``Joint Beam Training and Data Transmission Control for Mmwave Delay-Sensitive Communications: A Parallel Reinforcement Learning Approach," IEEE J.\ of Selected Topics in Signal Processing, Vol.\ 16, pp.\ 447-459.

\ref[Lee20] Lee, E.\ H., 2020.\ ``Budget-Constrained Bayesian Optimization, Doctoral dissertation, Cornell University.

\ref[Man74] Mandl, P., 1974.\ ``Estimation and Control in Markov Chains," Advances in Applied Probability, Vol.\ 6, pp.\ 40-60.

\ref[MEJ19] Molin, A., Esen, H., and Johansson, K.\ H., 2019.\ ``Scheduling Networked State Estimators Based on Value of Information," Automatica, Vol.\ 110.

\ref[PSC22] Paulson, J.\ A., Sonouifar, F., and Chakrabarty, A., 2022.\ ``Efficient Multi-Step Lookahead Bayesian Optimization with Local Search Constraints," IEEE Conf.\ on Decision and Control, pp.\ 123-129.

\ref[PoF08] Powell, W.\ B., and Frazier, P., 2008.\ ``Optimal Learning," in State-of-the-Art Decision-Making Tools in the Information-Intensive Age, INFORMS, pp.\ 213-246.

\ref[PoR12] Powell, W.\ B., and Ryzhov, I.\ O., 2012.\ Optimal Learning, J.\ Wiley, N.\ Y.

\ref[QHS05] Queipo, N.\ V., Haftka, R.\ T., Shyy, W., Goel, T., Vaidyanathan, R., and Tucker, P.\ K., 2005.\ ``Surrogate-Based Analysis and Optimization," Progress in Aerospace Sciences, Vol.\ 41, pp.\ 1-28.
 
 \ref[RPF12] Ryzhov, I.\ O., Powell, W.\ B., and Frazier, P.\ I., 2012.\ ``The Knowledge Gradient Algorithm for a General Class of Online Learning Problems," Operations Research, Vol.\ 60, pp.\ 180-195.
 
\ref[RaW06] Rasmussen, C.\ E., and Williams, C.\ K., 2006.\ Gaussian Processes for Machine Learning,   MIT Press, Cambridge, MA.

\ref[SSW16] Shahriari, B., Swersky, K., Wang, Z., Adams, R.\ P., and De Freitas, N., 2015.\ ``Taking the Human Out of the Loop: A Review of Bayesian Optimization," Proc.\ of IEEE, Vol.\ 104, pp.\ 148-175.

\ref[SWM89] Sacks, J., Welch, W.\ J., Mitchell, T.\ J., and Wynn, H.\ P., 1989.\ ``Design and Analysis of Computer Experiments," Statistical Science, Vol.\ 4, pp.\ 409-423.

\ref[Sas02] Sasena, M.\ J., 2002.\ Flexibility and Efficiency Enhancements for Constrained Global Design Optimization with Kriging Approximations, Ph.D.\ Thesis, Univ.\ of Michigan.

\ref [Sel22] Selby A., 2022.\ ``The Best Strategies for Wordle (last edited on 17 March 2022)," Available at URL (Accessed: 14 November 2022).

\ref[YuK20] Yue, X., and Kontar, R.\ A., 2020.\ ``Lookahead Bayesian Optimization via Rollout: Guarantees and Sequential Rolling Horizons," arXiv preprint arXiv:1911.01004.

\end